\pdfoutput=1

\documentclass[11pt]{article}
\usepackage{colortbl}
\usepackage[table]{xcolor}
\usepackage[preprint]{acl}

\usepackage{times}
\usepackage{latexsym}

\usepackage[T1]{fontenc}
\usepackage[utf8]{inputenc}
\usepackage{microtype}
\usepackage{inconsolata}

\usepackage{siunitx}
\usepackage{graphicx}
\usepackage{eso-pic}
\usepackage[absolute,overlay]{textpos}

\usepackage{subcaption}
\usepackage{times}
\usepackage{latexsym}
\usepackage[T1]{fontenc}
\usepackage[utf8]{inputenc}
\usepackage{pifont}

\usepackage{soul}
\usepackage{microtype}
\usepackage{tabularx}
\usepackage{xspace}
\usepackage{makecell}
\usepackage{tipa}
\usepackage{graphicx}
\usepackage{booktabs}
\usepackage{multirow}
\usepackage{caption}
\usepackage{amsmath} 
\usepackage{arydshln}
\usepackage{amssymb}  
\usepackage{xcolor}   
\usepackage{tcolorbox}
\tcbuselibrary{breakable,listings,skins}
\newcounter{promptbox}

\newtcolorbox{promptbox}[2][]{
  breakable,
  colback=white,
  colframe=black,
  fonttitle=\bfseries,
  title={#2},
  #1
}

\usepackage{tabularx}
\usepackage{color}
\usepackage{booktabs}
\usepackage{arydshln}

\usepackage[normalem]{ulem}
\definecolor{BrightOrange}{HTML}{FF6347}  
\definecolor{MutedBlue}{HTML}{5091C8}
\definecolor{BrightGreen}{HTML}{32CD32}    
\definecolor{BrightPurple}{HTML}{9370DB}  

\title{AnchorMem: Anchored Facts with Associative Contexts for Building Memory in Large Language Models}

\author{
  Zhanyu Shen\textsuperscript{1,2}, 
  Sijie Cheng\textsuperscript{2,3}\thanks{Corresponding authors.}, 
  Zhicheng Guo\textsuperscript{2,3}, Weiqin Wang\textsuperscript{1} \\
  {\bf Yile Wang\textsuperscript{1}\footnotemark[1], }
  {\bf Hui Huang\textsuperscript{1}}  \\
  \textsuperscript{1}College of Computer Science and Software Engineering, Shenzhen University\\
  \textsuperscript{2}RayNeo.AI  
  \textsuperscript{3}Tsinghua University\\
  \texttt{zhanyushen.s@gmail.com, csj23@mails.tsinghua.edu.cn, wangyile@szu.edu.cn} 
}

\begin{document}

\begin{textblock*}{5cm}(2.2cm,1cm) 
    \includegraphics[width=3cm]{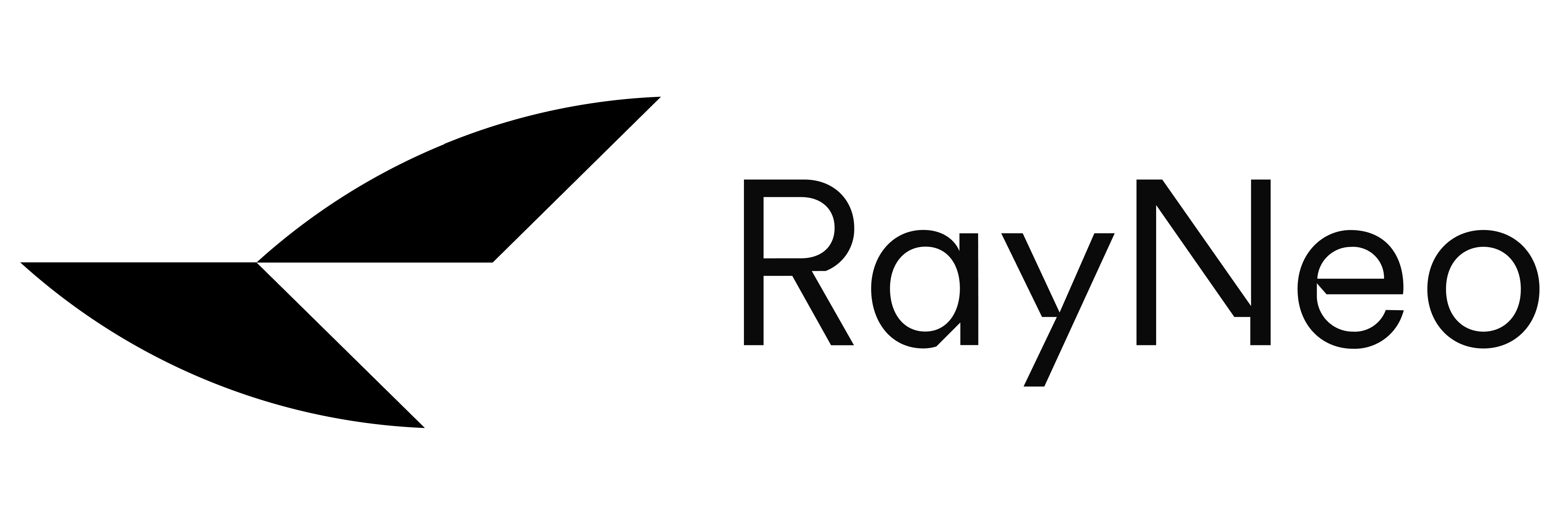}
\end{textblock*}

\maketitle

\begin{abstract}
While large language models have achieved remarkable performance in complex tasks, they still need a memory system to utilize historical experience in long-term interactions. Existing memory methods (e.g., A-Mem, Mem0) place excessive emphasis on organizing interactions by frequently rewriting them, however, this heavy reliance on summarization risks diluting essential contextual nuances and obscuring key retrieval features. To bridge this gap, we introduce AnchorMem, a novel memory framework inspired by the Proust Phenomenon in cognitive science, where a specific anchor triggers a holistic recollection. We propose a method that decouples the retrieval unit from the generation context. AnchorMem extracts atomic facts from interaction history to serve as retrieval anchors, while preserving the original context as the immutable context. To reveal implicit narrative cues, we construct an associative event graph that uses higher-order event links that bind sets of related facts into shared event representations, strengthening cross-memory integration without relying on generic entities as bridges. During retrieval, the system anchors queries to specific facts and events to locate relevant memories, but reconstructs the context using the associated raw chunks and  events. Our method reconciles fine-grained retrieval with the contextual integrity of interactions. Experiments across three closed-source and open-source models on the LoCoMo benchmark demonstrate that AnchorMem significantly outperforms baselines. Code is available at \href{https://github.com/RayNeo-AI-2025/AnchorMem}{https://github.com/RayNeo-AI-2025/AnchorMem}.
\end{abstract}

\section{Introduction}
Although large language models (LLMs;~\citealp{achiam2023gpt-intro, 2025deepseek-intro}) have demonstrated remarkable capabilities in single-turn inference and generation tasks, their stateless nature remains a primary bottleneck hindering their application in long-term, complex interactions~\cite{wu2025longmemeval-intro}. While extending context windows can partially mitigate context loss, this approach incurs substantial computational costs and leaves LLMs susceptible to the \emph{Lost-in-the-Middle} phenomenon during long-term, multi-turn interactions~\cite{liu2024lost-into, laban2025llms-into}. Memory systems enable LLMs to aggregate and synthesize relevant cues in a training-free manner, facilitating continual adaptation during interactions with the environment. Such capabilities are paramount for maintaining consistency across prolonged interactions or complex tasks~\cite{kang-etal-2025-memoryos}.

\begin{figure*}[t!]
    \centering
    \includegraphics[scale=1.16]{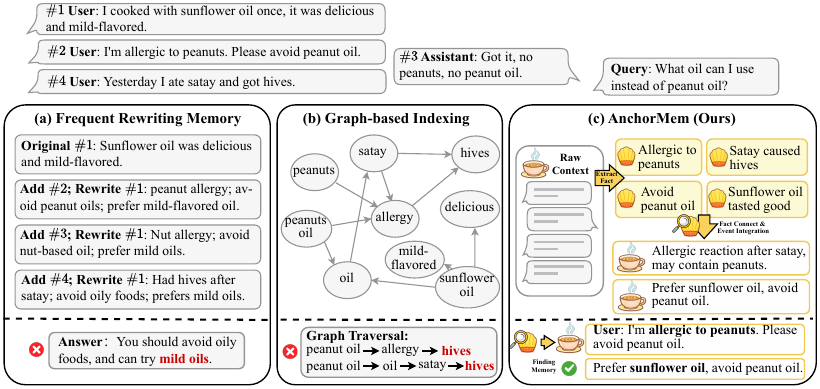}
    \caption{Comparison of memory paradigms. (a) Frequent rewriting overwrites context, (b) graph-based indexing relies on graph traversal with noisy links, while (c) AnchorMem anchors retrieval on atomic facts and events and reconstructs responses using the associated original context, preserving contextual integrity. \includegraphics[width=0.015\textwidth]{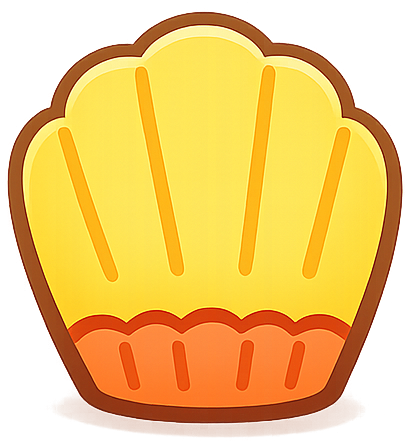} represents an atomic fact. \includegraphics[width=0.02\textwidth]{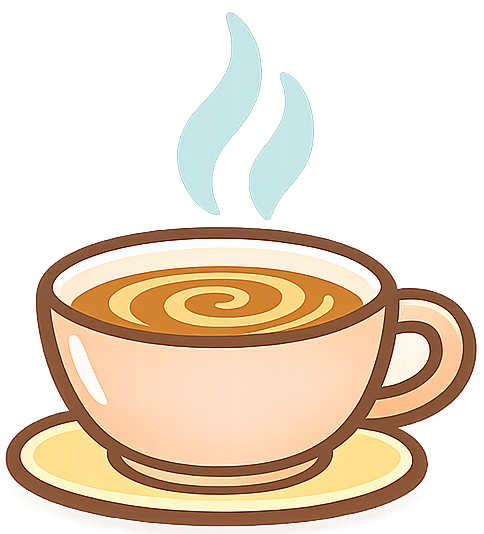} represents a related memory.}
    \label{figure:firt_figure}
\end{figure*}

Recent work, motivated by principles from human cognition~\citep{RW-MS-fang2025lightmem, RW-GR-gutierrez2025from} and knowledge management~\citep{RW-MS-xu2025amem, kang-etal-2025-memoryos}, increasingly designs memory for LLMs in a hierarchical manner, separating storage, consolidation, and retrieval to support long-horizon behavior. However, these approaches encounter significant limitations in maintaining contextual integrity and retrieval precision during dynamic interactions. Generative memory systems, such as A-Mem~\citep{RW-MS-xu2025amem} and Mem0~\citep{chhikara2025mem0}, typically employ a ``retrieve-and-rewrite'' mechanism to consolidate new information. As illustrated in Figure~\ref{figure:firt_figure} (a), this frequent rewriting effectively compresses history but may inadvertently result in contextual blurring, where specific nuances of the original context are diluted or lost through repetitive summarization. Alternatively, graph-based indexing strategies~\citep{RW-GR-edge2024graphrag, RW-GR-gutierrez2025from} excel at structuring heterogeneous knowledge sources by explicitly modeling entity relationships. However, within interaction data, this method risks establishing spurious connections driven by high-frequency but semantically generic entities. As shown in Figure~\ref{figure:firt_figure} (b), common terms (e.g., ``oil'') may act as misleading bridges, guiding the retrieval algorithm to traverse incorrect paths and conflate unrelated events.

To address these issues, we propose AnchorMem, a framework inspired by the cognitive theory of Involuntary Autobiographical Memory (IAM;~\citealp{berntsen2010unbidden-bio, berntsen2021involuntary-bio}). Often popularized as the Proust Phenomenon, where the specific taste of a madeleine involuntarily evokes a vivid, volumetric recollection of the past, this concept suggests that precise, concrete cues can effectively index complex episodic experiences~\citep{conway2000construction-bio}. Mimicking this mechanism, AnchorMem decouples the retrieval unit from the generation context. We extract discrete atomic facts from history to serve as retrieval anchors, while retaining the associated raw context as the immutable memory content, as shown in Figure~\ref{figure:firt_figure} (c). This design mitigates the aforementioned limitations. By anchoring on specific facts, we avoid the spurious connections caused by generic entities in graph indexing. Furthermore, by constructing a global associative event graph, we efficiently establish logical connections across distinct memories. This strategy allows for the integration of fragmented information without the need for frequent, destructive rewriting.

We evaluate AnchorMem on the long-term interactions benchmark LoCoMo~\citep{maharana-etal-2024-evaluating-locomo}. Extensive experiments across three open-source and closed-source models demonstrate that our approach achieves significant performance improvements under various evaluation metrics, markedly surpassing existing baselines such as the memory system LightMem~\citep{RW-MS-fang2025lightmem} and the RAG-based HippoRAG 2~\citep{RW-GR-gutierrez2025from}. Meanwhile, by circumventing the need for frequent rewriting, AnchorMem achieves the fastest memory construction speed among all compared methods while maintaining a balanced token consumption.

\section{Related Work}
\noindent\textbf{Generative Memory Systems.}
Memory systems aim to simulate human cognition by integrating historical interactions into an external store for long-horizon coherence and personalization~\cite{RW-MS-10.1145/3748302,RW-MS-mei2025survey}. Early explorations such as MemGPT~\cite{packer2023memgpt-intro} and MemoryBank~\cite{RW-MS-zhong2024memorybank} primarily extended limited context windows via summarization, retrieval, and time-aware decay mechanisms, but often lacked robust knowledge fusion across turns and topics. Recent work emphasizes more structured organization and continual evolution. A-Mem~\cite{RW-MS-xu2025amem} transforms interactions into structured notes and uses retrieval to drive note linkage and evolution, while Mem0~\cite{chhikara2025mem0} performs fine-grained extraction and state updates over specific interaction fragments. Inspired by system-level memory management, MemoryOS~\cite{kang-etal-2025-memoryos} organizes memories into a three-tier hierarchy (short-, mid-, long-term) and coordinates storage, updating, retrieval, and generation;  LightMem~\cite{RW-MS-fang2025lightmem} filters redundant interaction content via lightweight compression, then groups the remaining turns by topic and produces topic-wise summaries for memory construction. 

Mainstream generative memory systems rely on LLM-driven rewriting, which can blur contextual nuances and introduces additional maintenance overhead during long interactions. In contrast, our method avoids frequent rewriting by using atomic facts as retrieval anchors while preserving the original context as immutable generation context.

\noindent\textbf{Pre-Retrieval Indexing.}
Retrieval-Augmented Generation (RAG) enhances LLMs by retrieving external text evidence to ground generation~\cite{lewis2020retrieval, gao2023retrieval}. Standard RAG splits long texts into fixed-length chunks and embeds them for similarity search, however, when the chunk itself simultaneously serves as the retrieval unit and the generation context, segmentation inevitably cuts through semantically related information~\cite{RW-CR-duarte-etal-2024-lumberchunker, RW-CR-yu-etal-2024-chain, RW-CR-qu2025semantic}. To mitigate fragmentation, graph-based indexing constructs explicit structures (e.g., entity-relation graphs) to connect information units before retrieval. GraphRAG builds an entity knowledge graph and derives higher-level summaries from detected communities~\cite{RW-GR-edge2024graphrag}, while LightRAG simplifies this pipeline and retrieves from both low- and high-level graph views for efficiency~\cite{guo-etal-2025-lightrag}. HippoRAG 2~\cite{RW-GR-gutierrez2025from} uses an entity graph mainly as an associativity signal to rerank dense retrieval results.

Pre-retrieval indexing based on chunking can lose narrative continuity, while entity-centric triples are often context-poor for interaction memories and may introduce spurious links. AnchorMem anchors retrieval on atomic facts but reconstructs context from the associated raw context and events.

\section{Preliminaries}
Let the interaction content be denoted as $\mathcal{C} = \{C_1, C_2, \dots, C_T\}$, where each $C_i$ represents a discrete unit of interaction content. We formalize three paradigms of memory construction.

\noindent\textbf{Definition 1: Frequent Rewriting Memory.}
This paradigm processes interactions sequentially to maintain a discrete set of memories $\mathcal{M}$. At time step $t$, a tentative memory unit $M^*_t$ is extracted from the content $C_t$. The system then retrieves a subset of relevant existing memories $M_{ret} \subset \mathcal{M}_{t-1}$. An LLM-based function $\mathcal{R}$ determines how to rewrite the memories and generates:
\begin{equation}
    \mathcal{M}_t = (\mathcal{M}_{t-1} \setminus M_{ret}) \cup \mathcal{R}(M^*_t, M_{ret}).
\end{equation}
Here, $\mathcal{R}$ outputs the updated set of memories, potentially modifying or merging $M^*_t$ and elements of $M_{ret}$.

\noindent\textbf{Definition 2: Graph-based Indexing.}
This approach structures knowledge by extracting entities ($e$) and semantic relations ($r_{kg}$) from the content $x$. The knowledge graph constructed as:
\begin{equation}
    \mathcal{G}_{KG} = \bigcup_{x \in \mathcal{C}} \left( \text{Extract}(x) \cup \{ (e, r_{src}, x^*) \} \right),
\end{equation}
where $\text{Extract}(x)$ produces triplets $\{ (e_i, r_{kg}, e_j) \}$. The tuple $(e, r_{src}, x^*)$ denotes an optional provenance edge linking entities back to their source node $x^*$.

\begin{figure*}[t!]
    \centering
    \includegraphics[scale=1.16]{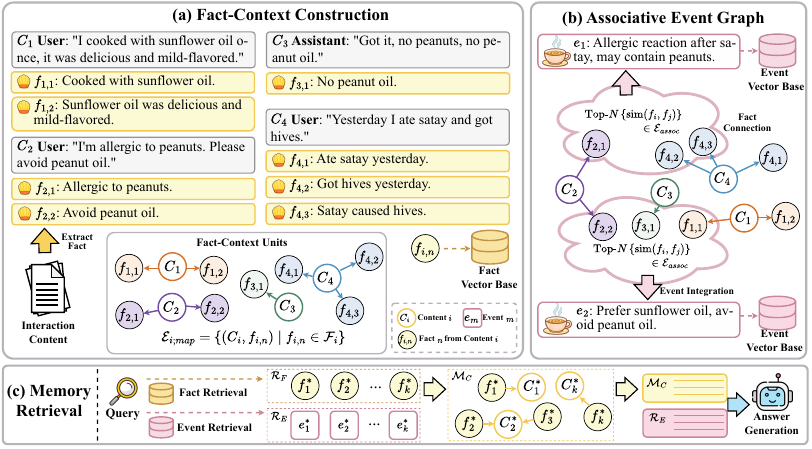}
    \caption{The overview of the AnchorMem framework. (a) Fact-Context Construction extracts atomic facts from interactions to serve as retrieval anchors for the immutable context. (b) Associative Event Graph integrates fragmented context into associative events. (c) Memory Retrieval constructs the memory using anchored facts and associative events for generation.}
    \label{figure:method}
\end{figure*}

\noindent\textbf{Definition 3: AnchorMem.}
We formulate AnchorMem as a heterogeneous graph $\mathcal{G} = (\mathcal{V}, \mathcal{E})$, where the node set comprises Interaction Contexts $\mathcal{V}_C$, Atomic Facts $\mathcal{V}_F$, and Associative Events $\mathcal{V}_E$:
\begin{equation}
\mathcal{V} = \mathcal{V}_C \cup \mathcal{V}_F \cup \mathcal{V}_E.
\end{equation}
The edge set $\mathcal{E}$ is composed of mapping and similarity relations:
\begin{equation}
\mathcal{E} = \mathcal{E}_{map} \cup \mathcal{E}_{assoc},
\end{equation}
where $\mathcal{E}_{map}$ represents a surjective mapping linking atomic facts to their immutable source contexts ($\mathcal{V}_F \to \mathcal{V}_C$), and $\mathcal{E}_{assoc}$ denotes semantic connections between atomic facts ($\mathcal{V}_F \times \mathcal{V}_F$). This formulation reconciles the granularity of atomic retrieval with the fidelity of the raw interaction context.

\section{Method}

Drawing on the Proust Moment, we propose AnchorMem, a memory framework designed to reconcile the specificity of atomic facts with the contextual integrity required for long-term interaction. We construct interaction contents into Fact-Context Units (FCUs; §\ref{sec:FCU}). These units form the foundation of the Associative Event Graph (AEG; §\ref{sec:AEG}), allowing retrieval to anchor on specific details while reconstructing the narrative volume. Our retrieval mechanism (§\ref{sec:retrieval}) leverages these structures to anchor queries on specific facts or events, subsequently recovering the complete, immutable context for generation.
\subsection{Fact-Context Construction}
\label{sec:FCU}

The foundation of AnchorMem lies in the construction of Fact-Context Units, which decouple storage into atomic retrieval units and immutable narrative contexts, as shown in Figure~\ref{figure:method} (a). Unlike methods that rewrite history into summaries, we maintain the interaction content $C_i \in \mathcal{V}_C$ as the ground truth and extract atomic facts to serve as searchable indices.

For each incoming interaction content $C_i$, we derive a set of atomic facts $\mathcal{F}_i$ by prompting the LLM $\pi$ with an extraction prompt $p_{ext}$ (Appendix~\ref{prompt:ext_fact}): 
\begin{equation}
\mathcal{F}_i = \pi(p_{ext}, C_i) = \{f_{i,1}, f_{i,2}, \dots, f_{i,n}\},
\end{equation}
where each $f_{i,n}$ is a concise statement representing a specific detail derived from the interaction context. We then update the global set of fact nodes $\mathcal{V}_F \leftarrow \mathcal{V}_F \cup \mathcal{F}_i$. We define a surjective mapping $\mathcal{M}(f_{i,n}) = C_i$ that binds each extracted fact $f \in \mathcal{F}_i$ to the current context $C_i$. This mapping formally constitutes the mapping edges in our graph, denoted as:
\begin{equation}
\mathcal{E}_{map} = \{ (\mathcal{M}(f), f) \mid f \in \mathcal{V}_F \}.
\end{equation}
This mapping ensures that any retrieved fact acts as an anchor, immediately directing the system to the original, immutable context.

\begin{table*}[t!]
    \centering
    \renewcommand{\arraystretch}{1.2} 
    \setlength{\extrarowheight}{0pt}
    \newcommand{\mc}[1]{\multicolumn{1}{c}{#1}}
    \scalebox{0.61}{
    \begin{tabular}{l  ccc  ccc  ccc  ccc  ccc}
        \toprule
        \multicolumn{1}{c}{\multirow{2}{*}{\textbf{Method}}}  & \multicolumn{3}{c}{\textbf{Single Hop}} & \multicolumn{3}{c}{\textbf{Multi Hop}} & \multicolumn{3}{c}{\textbf{Open Domain}} & \multicolumn{3}{c}{\textbf{Temporal}} & \multicolumn{3}{c}{\textbf{Average}}\\
        \cmidrule(lr){2-4} \cmidrule(lr){5-7} \cmidrule(lr){8-10} \cmidrule(lr){11-13} \cmidrule{13-16}
        \multicolumn{1}{c}{} & \mc{F1} & \mc{BLEU} & \mc{ACC} & \mc{F1} & \mc{BLEU} & \mc{ACC} & \mc{F1} & \mc{BLEU} & \mc{ACC} & \mc{F1} & \mc{BLEU} & \mc{ACC} & \mc{F1} & \mc{BLEU} & \mc{ACC} \\

        \midrule
        \multicolumn{16}{c}{\textbf{GPT-4o-mini}} \\
        \midrule

        NaiveRAG &40.20&31.56&57.55 &20.41&14.81&39.36 &10.91&10.01&34.38 &32.07&25.56&34.27 &33.06&25.90&47.92 \\
        HippoRAG 2~\cite{RW-GR-gutierrez2025from} &46.08&32.66&66.11 &30.65&21.58&55.67 &24.87&17.11&50.00 &51.73&36.75&59.50  &43.11&30.52&61.82\\
        Mem0$^{*}$~\cite{chhikara2025mem0}  &47.65&38.72&72.93 &\textbf{38.72}&\underline{27.13}&\textbf{67.13} &\textbf{28.64}&\textbf{21.58}&\textbf{51.15} &48.93&40.51&55.51 &45.10&35.90&66.88 \\
        Mem0~\cite{chhikara2025mem0}  &29.69&24.59&45.54 &20.33&13.28&30.85 &14.90&10.22&28.12 &39.38&32.85&33.96  &29.07&23.34&39.31\\
        A-Mem~\cite{RW-MS-xu2025amem} &35.84&29.22&57.40 &23.19&16.71&42.91 &9.06&8.34&23.96 &38.40&33.72&51.09 &32.39&26.57&51.35\\
        MemoryOS$^{*}$~\cite{kang-etal-2025-memoryos} &48.62&42.99&- &35.27&25.22&- &20.02&16.52&- &41.15&30.76&- &42.84&35.53&- \\
        MemoryOS~\cite{kang-etal-2025-memoryos} &38.53&33.90&55.65 &35.89 &\textbf{27.41} &56.74 &22.95&17.85&42.71 &22.46&20.84&23.36 &33.73&28.99&48.31\\
        LightMem~\cite{RW-MS-fang2025lightmem}  &44.48&35.49&67.18 &29.04&20.17&50.00 &23.81&\underline{18.09}&41.67 &50.92&38.06&64.80  &41.71&32.14&61.95\\

        \hdashline
        \textbf{AnchorMem (Ours)} &\underline{55.84} &\underline{44.38} &\underline{78.60} &34.45&23.87&56.03 &25.38 &17.91 &47.92 &\underline{55.09} &\underline{43.79} &\textbf{68.85} &\underline{49.87} &\underline{38.85} &\underline{70.52} \\
        \textbf{AnchorMem (Ours)}$^{\dagger}$ &\textbf{58.11}&\textbf{46.26}&\textbf{82.16} &\underline{36.79}&24.57&\underline{63.83} &\underline{26.34}&16.72&\underline{51.04} &\textbf{56.60}&\textbf{45.34}&\underline{67.60} &\textbf{51.91}&\textbf{40.25}&\textbf{73.83} \\
        \midrule
 
        \multicolumn{16}{c}{\textbf{Qwen2.5-32B-Instruct}} \\
        \midrule
        
        NaiveRAG &\underline{43.12} &\underline{38.34} &59.57 &23.89 &17.38 &42.20 &14.19 &11.84 &34.38 &30.75 &25.36 &33.33 &35.16&30.09&49.35\\

        HippoRAG 2~\cite{RW-GR-gutierrez2025from} &38.11&26.13&62.66 &27.83&20.29&56.38 &\underline{23.26}&16.02&42.71 &\underline{43.64}&\underline{32.52}&56.07 &36.45&25.76&58.90 \\
        
        Mem0~\cite{chhikara2025mem0}  &24.30 &20.42 &36.75 &17.51 &12.46 &29.43 &14.81 &11.48&31.25 &32.80 &27.20 &31.15 &24.24&19.81&33.88\\
        A-Mem~\cite{RW-MS-xu2025amem}  &35.34&31.39&47.56 &25.56&17.98&43.62 &11.96&10.79&28.12 &29.72&25.37&34.58 &30.92&26.40&42.92\\
        MemoryOS~\cite{kang-etal-2025-memoryos} &38.53&33.90&55.65 &\textbf{35.89}&\textbf{27.41}&\underline{56.74} &22.95&\underline{17.85}&\underline{42.71} &24.46&20.84&23.36  &34.14&28.99&48.31\\
        
        LightMem~\cite{RW-MS-fang2025lightmem}  &42.55 &35.42 &\underline{64.33} &29.71 &23.40 &50.71 &15.58 &12.06 &40.63 &41.23 &31.40 &\underline{61.99}  &\underline{38.24}&\underline{30.93}&\underline{59.87}\\

        \hdashline
        \textbf{AnchorMem (Ours)} &\textbf{57.82}&\textbf{49.76}&\textbf{78.83} &\underline{35.53}&\textbf{25.71}&\textbf{58.51} &\textbf{25.62}&\textbf{19.95}&\textbf{44.79} &\textbf{53.12}&\textbf{45.26}&\textbf{62.31} &\textbf{50.57}&\textbf{42.54}&\textbf{69.54} \\
        
        \midrule

        \multicolumn{16}{c}{\textbf{Qwen2.5-7B-Instruct}} \\
        \midrule

        NaiveRAG  &40.43&35.27&59.45 &23.76&18.15&45.04 &15.67&13.17&42.71 &27.41&23.83&25.55  &33.12&28.37&48.70\\
        HippoRAG 2~\cite{RW-GR-gutierrez2025from} &27.00&18.34&67.78 &19.30&15.32&\underline{54.96} &16.70&13.14&\textbf{53.13} &26.04&18.40&\textbf{42.68} &24.75&17.48&\underline{59.26} \\
        
        Mem0~\cite{chhikara2025mem0}  &27.03&23.31&37.11 &18.69&13.92&30.85 &12.89&9.94&26.04 &25.37&20.35&21.81  &24.25 &20.11 &32.05\\
        A-Mem~\cite{RW-MS-xu2025amem}  &38.93&33.95&58.33 &23.70&16.62&46.81 &15.34&12.71&\underline{38.54} &32.87&27.55&28.66  &33.41&28.12&48.80\\
        
        MemoryOS~\cite{kang-etal-2025-memoryos}   &32.62&28.17&51.72 &25.94&19.10&40.07 &\underline{18.94}&14.81&35.42 &25.87&21.60&17.45 &29.14&24.31&41.43 \\
        
        LightMem~\cite{RW-MS-fang2025lightmem}  &\underline{41.34} &\underline{35.30} &59.93 &\underline{28.55} &\underline{21.90} &46.81 &\textbf{19.56} &\textbf{17.38} &34.38 &\underline{38.37} &\underline{28.26} &\underline{40.81}  &\underline{37.02} &\underline{30.26} &51.95\\
        
        \hdashline
        \textbf{AnchorMem (Ours)}  &\textbf{54.32} &\textbf{45.46} &\textbf{72.89} &\textbf{34.50} &\textbf{24.45} &\textbf{55.67} &18.40 &\underline{15.37} &\underline{38.54} &\textbf{43.85} &\textbf{31.65} &\underline{40.81}  &\textbf{46.27} &\textbf{36.86} &\textbf{60.91}\\
        \bottomrule
    \end{tabular}}
\caption{Performance comparison of various methods on the LoCoMo dataset. Among the related methods, the best results in each column are \textbf{in bold}, and the second-best results are \underline{underlined}. The $^*$ refers to the results reported in the original paper. The $^{\dagger}$ denotes a retrieval top-k of 20, consistent with Mem0$^{*}$, while all other settings remain unchanged.}
\label{tab:result_comparison}
\end{table*}

\subsection{Associative Event Graph}
\label{sec:AEG}

To capture high-level and cross-context relationships, we construct Associative Events nodes $\mathcal{V}_E$ by clustering atomic facts, as shown in Figure~\ref{figure:method} (b). While atomic facts provide specificity, events integrate fragmented details into coherent narratives. 

\noindent\textbf{Fact Connection.} We first identify latent connections by computing the pairwise cosine similarity between the embeddings $\mathbf{v}_f$ of all facts $f \in \mathcal{V}_F$. The affinity between two facts $f_i$ and $f_j$ implies a semantic edge in $\mathcal{E}_{assoc}$ defined as:
\begin{equation}
    \text{sim}(f_i, f_j) = \frac{\mathbf{v}_{f_i} \cdot \mathbf{v}_{f_j}}{\lVert \mathbf{v}_{f_i} \rVert \lVert \mathbf{v}_{f_j} \rVert}.
\end{equation}

To form a candidate cluster $\mathcal{N}_i$ for $f_i$, we retrieve the top-$N$ neighbors that satisfy the similarity constraint:
\begin{equation}
    \mathcal{N}_i = \mathop{\text{Top-}N}_{f_j \in \mathcal{V}_F \setminus \{f_i\}} \{ f_j \mid \text{sim}(f_i, f_j) > \tau \} \cup \{f_i\},
\end{equation}
where $\tau$ is an inter-fact similarity threshold.
To avoid excessive integration, we prune the candidate set by calculating the maximum overlapping content  $O_{max}$:
\begin{equation}
O_{max} = \max_{C \in \mathcal{V}_C} |\{ f \in \mathcal{N}_i \mid \mathcal{M}(f) = C \}|,
\end{equation}
If $O_{max} \ge \lceil N/2 \rceil$, the cluster is considered locally redundant and is discarded.

\noindent\textbf{Event Integration.} For the retained clusters, we construct the final event set $\mathcal{S}_i$ by pairing unique contexts with their most representative facts. We deduplicate $\mathcal{N}_i$ based on the mapped contexts $\mathcal{M}(f)$, ensuring each context $C$ appears only once:
\begin{equation}
\mathcal{S}_k = \{ (C, f) \mid C \in \{ \mathcal{M}(f) \mid f \in \mathcal{N}_i \} \}.
\end{equation}
Then, we employ an LLM with an integration prompt $p_{inte}$ (Appendix~\ref{prompt:inte_event}) to synthesize the event node $e_k$:
\begin{equation}
    e_k = \pi( p_{inte}, \mathcal{S}_k).
\end{equation}
The associative event $e_k$ summarizes the relationship between the scattered facts, serving as a high-level retrieval anchor.

\begin{table*}[t!]
    \centering
    \scalebox{0.78}{
    \begin{tabular}{lcccccc}
        \toprule
       \textbf{Method}& \textbf{ACC$\uparrow$} &\textbf{Summary (k)$\downarrow$} & \textbf{Update (k)$\downarrow$} & \textbf{Total (k)$\downarrow$} & \textbf{Calls$\downarrow$} & \textbf{Runtime (s)$\downarrow$}\\
       \midrule
        HippoRAG 2~\cite{RW-GR-gutierrez2025from} &59.48 &4,880.35 &4,673.47 &9,553.83 &12,760 &5,410.16\\
        Mem0~\cite{chhikara2025mem0} &33.88 &5,902.29 &8,542.92 &14,445.20 &10,147 &6,200.04\\
        A-Mem~\cite{RW-MS-xu2025amem} &42.92 &2,048.29 &12,527.00 &14,575.30 &11,763 &99,252.15 \\
        MemoryOS~\cite{kang-etal-2025-memoryos} &48.31 &1,762.96 &2,462.42 &4,225.38 &14,318 &28,262.44 \\
        LightMem~\cite{RW-MS-fang2025lightmem} &59.87 &951.12 &178.50 &1,129.62 &538 &8,144.30 \\
        
        \hdashline
        \textbf{AnchorMem (Ours)} &69.54 &2,340.35 &5,129.10 &7,469.45 &6,577 &5,073.75 \\
        
        \bottomrule
    \end{tabular}}
\caption{Cost-efficiency comparison of various methods on the LoCoMo dataset with Qwen2.5-32B-Instruct. We report Accuracy (ACC), Token consumption (Summary, Update, and Total), Call Counts, and Runtime. Token usage is measured in thousands (k), and Runtime in seconds (s). The symbols $\uparrow$ and $\downarrow$ indicate that higher and lower values are better, respectively.}
\label{tab:efficiency_comparison}
\end{table*}

\subsection{Memory Retrieval}
\label{sec:retrieval}

Our retrieval strategy capitalizes on the explicit associative structures constructed during the memory indexing phase. Instead of performing operationally heavier graph traversals, we directly access relevant memory through vector similarity, as shown in Figure~\ref{figure:method} (c). Given a query $q$, 
we first compute the similarity between its embedding $\mathbf{v}_q$ and the embeddings of facts $\mathbf{v}_f$ and events $\mathbf{v}_e$ to identify the top-$k$ most relevant candidates:
\begin{align}
    \mathcal{R}_{F} = \mathop{\text{Top-}k}_{f \in \mathcal{V}_F}(\text{sim}(\mathbf{v}_q, \mathbf{v}_f)), \\
    \mathcal{R}_{E} = \mathop{\text{Top-}k}_{e \in \mathcal{V}_E}(\text{sim}(\mathbf{v}_q, \mathbf{v}_e)).
\end{align}
Instead of using the facts $\mathcal{R}_{F}$ directly as context, we use them as anchors to traverse the edge set $\mathcal{E}_{map}$ and retrieve the original interaction contents. This restores the full narrative fidelity:
\begin{equation}
    \mathcal{M}_{C} = \{ C^* \in \mathcal{V}_C \mid \exists f^* \in \mathcal{R}_{F}, (f^*, C^*) \in \mathcal{E}_{map} \},
\end{equation}
where the $^*$ denotes retrieved nodes used for the final generation. The final memory context $\mathcal{M}_{final}$ combines the exact historical nuances and the high-level event summaries:
\begin{equation}
    \mathcal{M}_{final} = \mathcal{M}_{C} \cup \mathcal{R}_{E}.
\end{equation}
By mapping facts back to $\mathcal{M}_{C}$, the model accesses the exact phrasing and nuance of the original context, avoiding the information loss typical of rewriting methods. $\mathcal{R}_{E}$ provides synthesized insights spanning multiple conversation turns, enabling the model to grasp complex, long-term dependencies that a single interaction cannot fully convey.

\section{Experiments}
\subsection{Settings}
\noindent\textbf{Dataset and Baseline Methods.}
To evaluate the effectiveness and efficiency of the proposed method, we conducted experiments on the LoCoMo~\cite{maharana-etal-2024-evaluating-locomo} dataset. Furthermore, we performed comparative experiments between AnchorMem against several other relevant baselines, namely Naive RAG, HippoRAG 2~\cite{RW-GR-gutierrez2025from}, Mem0~\cite{chhikara2025mem0}, A-Mem~\cite{RW-MS-xu2025amem}, MemoryOS~\cite{kang-etal-2025-memoryos}, and LightMem~\cite{RW-MS-fang2025lightmem}. Detailed descriptions of the datasets and baseline methods are provided in the Appendix~\ref{appendix:exp}.

\noindent\textbf{Implementation Details.}
For all comparative methods, three models were employed as LLM backbones: Qwen2.5-7B-Instruct and Qwen2.5-32B-Instruct~\cite{qwen2025qwen25technicalreport} were deployed locally via vLLM~\cite{kwon2023efficient-vllm}, while GPT-4o-mini was accessed through the official OpenAI API\footnote{\url{https://openai.com/api/}}. To ensure a fair comparison, we fixed the retrieval top-$k$ to 10 for all baselines and our proposed method. Furthermore, the all-MiniLM-L6-v2~\cite{reimers-2019-sentence-bert} embedding model was utilized across all experiments.

\noindent\textbf{Metrics.}
To evaluate performance, we report three metrics: F1 and BLEU-1 are employed to measure the overlap between generated responses and reference answers, ensuring faithfulness to the source text; additionally, Accuracy (ACC) is measured via an LLM-as-Judge to assess the comprehensiveness and logical coherence of the responses. The LLM Judge is based on a locally deployed Qwen2.5-32B-Instruct model served with vLLM~\cite{kwon2023efficient-vllm}, with temperature set to 0, detailed in Appendix~\ref{prompt:llm_judge}, where we also provide its stability analysis. To verify cost-efficiency, following the protocol of LightMem~\cite{RW-MS-fang2025lightmem}, we report LLM token consumption, call counts, and runtime during both the summary and update phases.

\subsection{Main Results}
\noindent\textbf{Performance.}
We conducted a comparative evaluation of five memory methods on the LoCoMo dataset, with the results presented in Table~\ref{tab:result_comparison}. The results demonstrate that AnchorMem consistently achieves superior performance across all evaluated metrics and large language model backbones, fully underscoring the effectiveness of utilizing facts as anchors to trigger memory content. On GPT-4o-mini, our method achieves an average F1 score of 49.87\%, surpassing the previous best-performing baseline, Mem0$^{*}$, which scores 45.10\%. This performance advantage is even more pronounced on open-source models, on Qwen2.5-32B-Instruct, AnchorMem reaches an average F1 of 50.57\%, outperforming the second-best approach, LightMem (38.24\%), by over 12\%. These results indicate that AnchorMem maintains robust adaptability and high performance. The outstanding F1 and BLEU-1 scores in these tasks indicate that our Fact-Context Units mechanism precisely targets crucial memory content, enabling the model to faithfully reconstruct original information when addressing factual queries.

\begin{table*}[t!]
    \centering
    \renewcommand{\arraystretch}{1.2} 
    \setlength{\extrarowheight}{0pt}
    \newcommand{\mc}[1]{\multicolumn{1}{c}{#1}}
    \scalebox{0.68}{
    \begin{tabular}{l ccc ccc ccc ccc ccc}
        \toprule
        \multicolumn{1}{c}{\multirow{2}{*}{\textbf{Method}}}  & \multicolumn{3}{c}{\textbf{Single Hop}} & \multicolumn{3}{c}{\textbf{Multi Hop}} & \multicolumn{3}{c}{\textbf{Open Domain}} & \multicolumn{3}{c}{\textbf{Temporal}} & \multicolumn{3}{c}{\textbf{Average}}\\
        
        \cmidrule(lr){2-4} \cmidrule(lr){5-7} \cmidrule(lr){8-10} \cmidrule(lr){11-13} \cmidrule{13-16}
        \multicolumn{1}{c}{} & \mc{F1} & \mc{BLEU} & \mc{ACC} & \mc{F1} & \mc{BLEU} & \mc{ACC} & \mc{F1} & \mc{BLEU} & \mc{ACC} & \mc{F1} & \mc{BLEU} & \mc{ACC} & \mc{F1} & \mc{BLEU} & \mc{ACC} \\
       \midrule
       \textbf{AnchorMem} &\textbf{57.82}&\textbf{49.76}&\textbf{78.83} &\textbf{35.53}&\textbf{25.71}&\textbf{58.51} &\textbf{25.62}&\textbf{19.95}&\textbf{44.79} &\textbf{53.12}&\textbf{45.26}&62.31 &\textbf{50.57}&\textbf{42.54}&\textbf{69.54} \\
        \hdashline
        w/o Context &24.03&20.36&35.20 &21.16&14.93&35.46 &16.95&15.57&27.08 &14.74&12.16&14.95 &21.13&17.36&30.52\\
        w/o FCU &48.88&42.19&67.65 &30.21&23.15&47.87 &21.77&17.53&41.67 &44.32&38.88&45.17 &42.79&36.48&57.73 \\
        w/o AEG &55.39&47.86&73.25 &30.49&22.25&51.06 &24.28&18.99&40.63 &51.87&44.17&\textbf{63.24} &48.16&40.60&65.06 \\
       
        \bottomrule
    \end{tabular}}
\caption{An ablation study was conducted to evaluate our proposed method against the Qwen2.5-32B-Instruct.}
\label{tab:module_ablation}
\end{table*}

\begin{figure*}[t]
  \centering
  \begin{subfigure}[t]{0.44\linewidth}
    \centering
    \includegraphics[width=\linewidth]{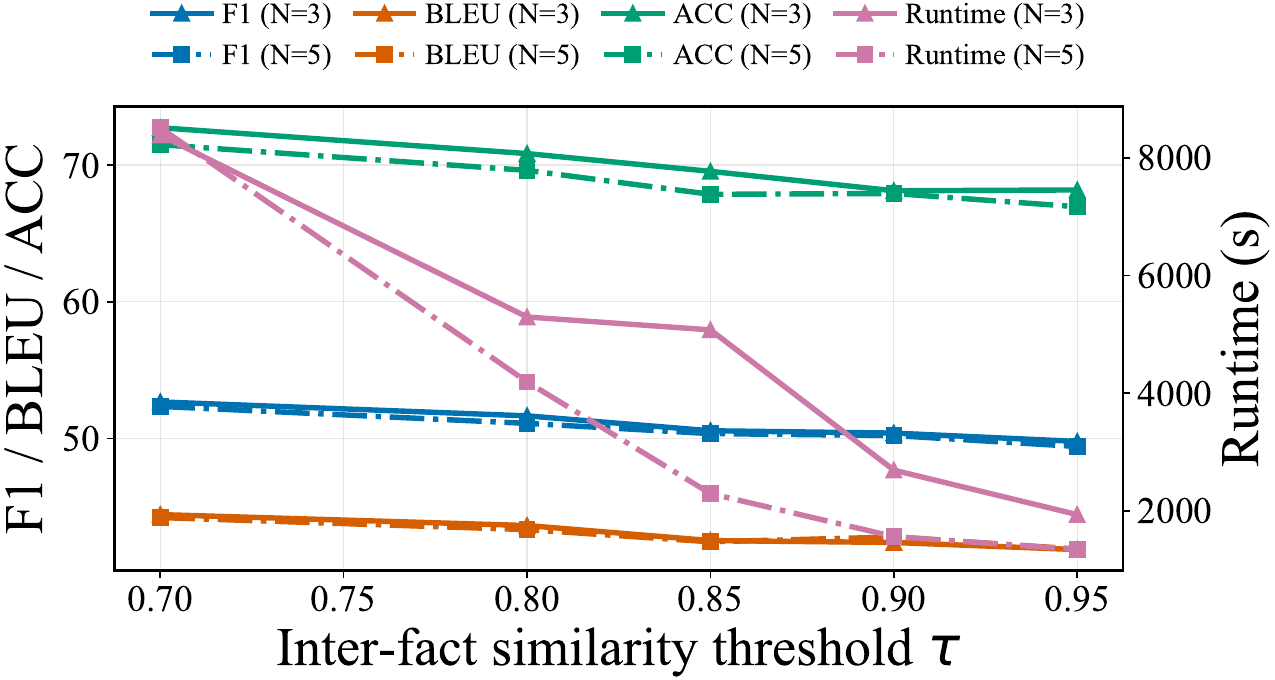}
    \caption{Performance and Runtime}
    \label{fig:aeg_hyparam_a}
  \end{subfigure}\hspace{0.03\linewidth}
  \begin{subfigure}[t]{0.44\linewidth}
    \centering
    \includegraphics[width=\linewidth]{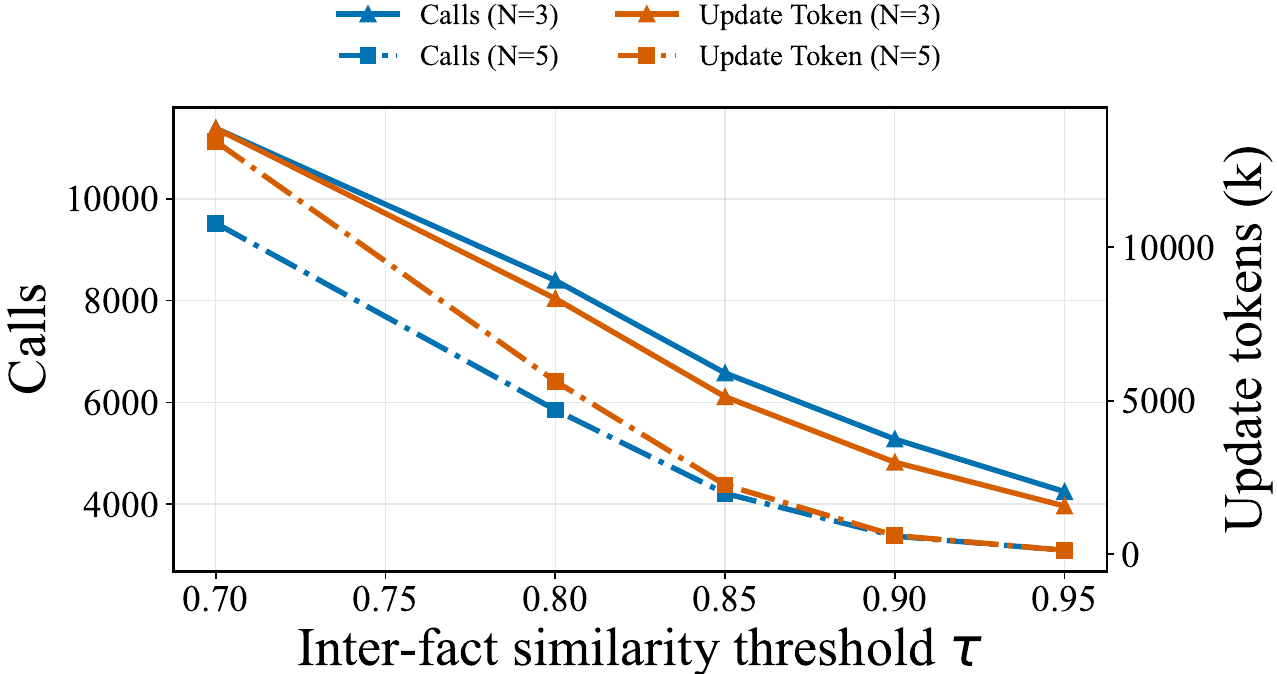}
    \caption{Calls and Update Tokens}
    \label{fig:aeg_hyparam_b}
  \end{subfigure}
  \caption{Impact of the Associative Event Graph hyperparameters, the inter-fact similarity threshold $\tau$ and the number of fact neighbors $N$, on performance and efficiency with Qwen2.5-32B-Instruct.}
  \label{fig:aeg_hyparam}
\end{figure*}

The results highlight distinct behavioral patterns between retrieval-based and memory-based approaches. 
NaiveRAG displays remarkable consistency across the three different LLM backbones, maintaining average F1 scores within a narrow range (33.06\% on GPT-4o-mini and 33.12\% on Qwen2.5-7B-Instruct), suggesting that model capacity exerts minimal influence on the performance of simple retrieval-augmented generation. In contrast, generative memory baselines generally exhibit performance variations that correlate with the capabilities of the underlying LLM backbones. While memory methods achieve competitive results in Multi Hop and Open Domain tasks through dynamic memory updates, their reliance on frequent summarization appears to compromise granular information retention. This is evidenced by their performance in Single Hop tasks on GPT-4o-mini, where MemoryOS (38.53\%) and LightMem (44.48\%) yield results comparable to NaiveRAG (40.20\%), likely due to the dilution of specific details during the rewriting process.

\noindent\textbf{Cost-Efficiency.}
Attributed to its robust compression and filtering strategies, LightMem remains the lightest-weight solution in terms of token consumption and API calls, as shown in Table~\ref{tab:efficiency_comparison}. However, our method, AnchorMem, achieves the highest accuracy (69.54\%) among the compared methods, significantly outperforming strong baselines such as LightMem (59.87\%) and HippoRAG 2 (59.48\%). By dispensing with the strategy of chronological memory addition, AnchorMem demonstrates remarkable efficiency in runtime, achieving the lowest execution time (5,073s) among all compared methods. It is approximately 1.6$\times$ faster than LightMem (8,144s) and significantly faster than MemoryOS and A-Mem. When compared to resource-intensive baselines like A-Mem and Mem0, AnchorMem proves superior across all dimensions, reducing total token usage by nearly 50\% meanwhile improving accuracy and runtime. Consequently, AnchorMem strikes a pragmatic balance, achieving optimal performance and speed at a reasonable token cost.

\begin{figure*}[t!]
  \centering
  
  \includegraphics[width=0.25\linewidth]{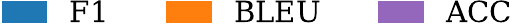} 
  \vspace{-0.2cm} 

  \vspace{0.2cm}
  \begin{subfigure}[b]{0.20\linewidth}
    \centering
    \includegraphics[width=\linewidth]{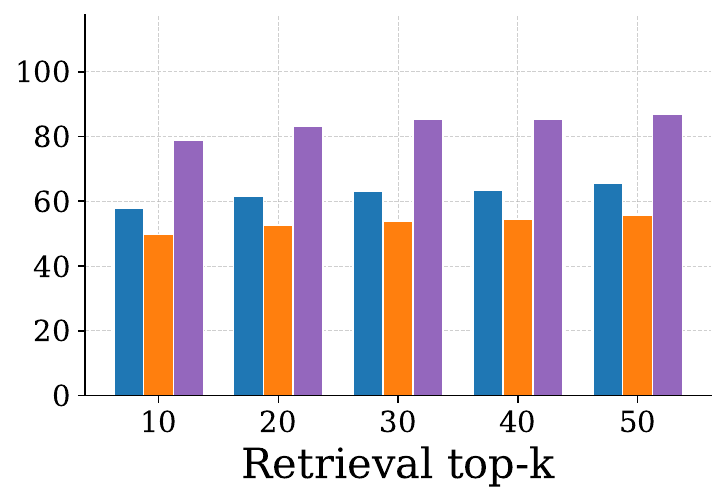}
    \caption{Single Hop}
    \label{fig:topk_hyparam_a}
  \end{subfigure}\hfill
  \begin{subfigure}[b]{0.20\linewidth}
    \centering
    \includegraphics[width=\linewidth]{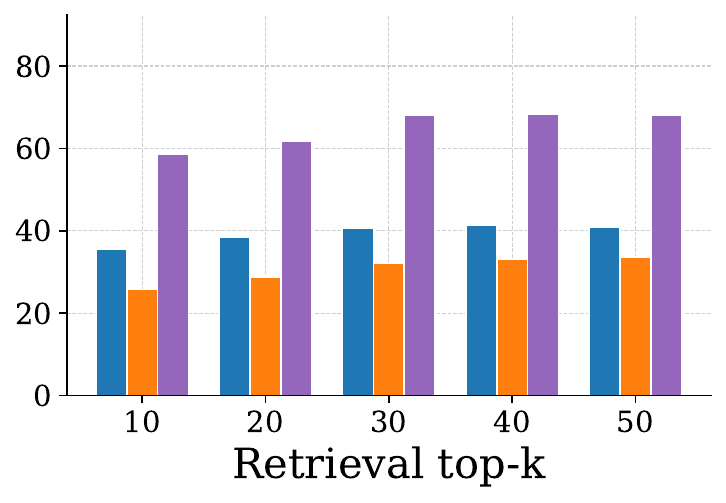}
    \caption{Multi Hop}
    \label{fig:topk_hyparam_b}
  \end{subfigure}\hfill
  \begin{subfigure}[b]{0.20\linewidth}
    \centering
    \includegraphics[width=\linewidth]{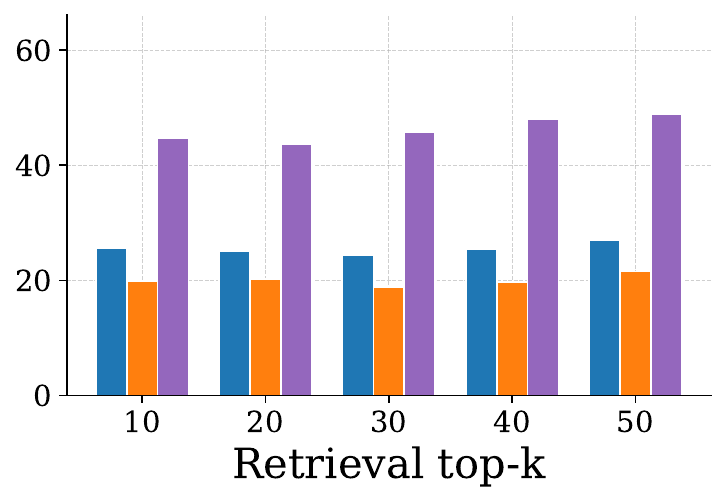}
    \caption{Open Domain}
    \label{fig:topk_hyparam_c}
  \end{subfigure}\hfill
  \begin{subfigure}[b]{0.20\linewidth}
    \centering
    \includegraphics[width=\linewidth]{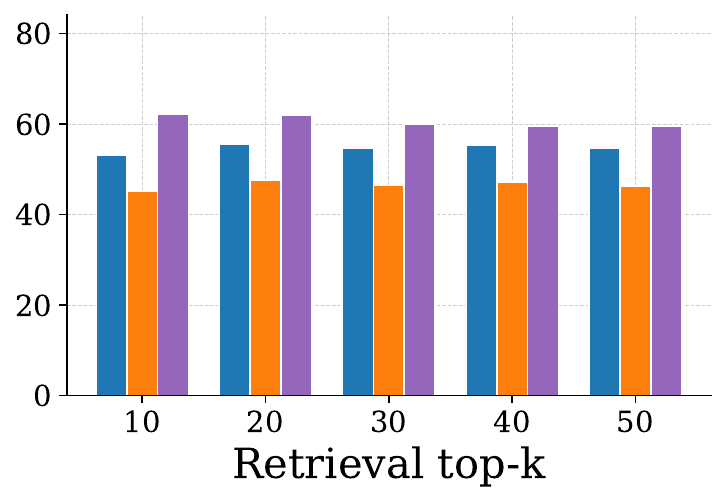}
    \caption{Temporal}
    \label{fig:topk_hyparam_d}
  \end{subfigure}\hfill
  \begin{subfigure}[b]{0.20\linewidth}
    \centering
    \includegraphics[width=\linewidth]{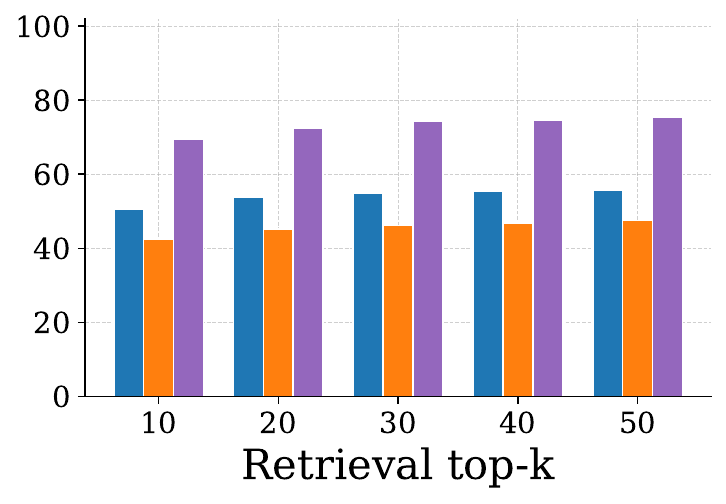}
    \caption{Average}
    \label{fig:topk_hyparam_e}
  \end{subfigure}

  \caption{Impact of memory retrieval parameter top-k across different task categories with Qwen2.5-32B-Instruct.}
  \label{fig:topk_hyparam}
\end{figure*}

\begin{table*}[t!]
    \centering
    \scalebox{0.78}{
    \begin{tabular}{llcccccc}
        \toprule
       \textbf{Phase}& \textbf{Component} &\textbf{Prompt (k)} & \textbf{Completion (k)} & \textbf{Total (k)} & \textbf{Calls} & \textbf{Total Time (s)} & \textbf{Avg. Time (s)}\\
       \midrule
       \multicolumn{1}{c}{\multirow{2}{*}{\textbf{Offline}}} & Fact Extraction & 2,022.02 & 318.33 & 2,340.35 & 3,007 & 1,263.89 &- \\
       & Event Integration & 4,558.31 & 570.79 & 5,129.10 & 3,570 & 3,809.85 &- \\
       \midrule
       \multicolumn{1}{c}{\multirow{2}{*}{\textbf{Online}}} & Vector Retrieval & - & - & - & - & 6.53 & 0.004 \\
       & Answer Generation & 2,917.92 & 12.69 & 2,930.60 & 1,540 & 870.68 & 0.57 \\
        \bottomrule
    \end{tabular}}
\caption{Phase-wise construction and inference cost on the LoCoMo dataset with Qwen2.5-32B-Instruct. We report Token consumption (Prompt, Completion, and Total), Call Counts, Total Runtime, and Average Runtime per query for online phase. Token usage is measured in thousands (k), and Runtime in seconds (s).}
\label{tab:component_efficiency_comparison}
\end{table*}

\begin{table}[t!]
    \centering
    \scalebox{0.85}{
    \begin{tabular}{lcc}
        \toprule
       \textbf{Statistic}& \textbf{Total} &\textbf{Average}\\
       \midrule
       Extracted Facts &16,023 &1,602\\
       Generated Events  &3,570 &357\\
       Discarded Clusters  &4,821 &482\\
        
        \bottomrule
    \end{tabular}}
\caption{Structural statistics of AEG construction on LoCoMo, reported as total counts and per-sample averages.}
\label{tab:structural_stat}
\end{table}

\vspace{-0.3em}
\section{Analyses}
\noindent\textbf{Ablation Study.}
To validate the AnchorMem architectural design, we conducted an ablation study to investigate the specific contributions of the original context, the Fact-Context Units (FCUs), and the Associative Event Graph (AEG) to model performance. As shown in Table~\ref{tab:module_ablation}, removing the original context (w/o Context) resulted in the most catastrophic performance decline, with average accuracy plummeting from 69.54\% to 30.52\%. Without the original context, which bears the full narrative, relying solely on fragmented fact segments causes the LLM to lose the narrative integrity and coherent logic, leading to a comprehensive collapse across all tasks.

Removing the Associative Event Graph (w/o AEG) resulted in a marked accuracy decline in Multi Hop tasks from 58.51\% to 51.06\%, validating its role in bridging cross-session reasoning. However, in Temporal tasks, the w/o AEG variant achieved a slightly higher accuracy (63.24\% vs 62.31\%), despite a lower F1 score (51.87\% vs 53.12\%). This suggests that the associative links effectively expand the search breadth to capture richer context. While this improves the overall answer quality, the broader scope may introduce minor noise during Temporal tasks compared to the restricted retrieval range of the variant without the AEG. Taken together, these results validate the synergistic effects of AnchorMem's design in balancing retrieval breadth with precision.

\noindent\textbf{Impact of AEG Hyperparameters.}
To determine the optimal configuration for the AEG, we analyzed the impact of the inter-fact similarity threshold $\tau$ and the number of fact neighbors $N$ on performance. As illustrated in Figure~\ref{fig:aeg_hyparam} and detailed in Appendix~\ref{appendix:hyper_aeg}, there is a distinct trade-off between semantic retention and computational efficiency. Lower $\tau$ values promote dense connectivity, benefiting accuracy and F1 scores but incurring higher runtime and token consumption due to the increased volume of fusion operations. Conversely, higher $\tau$ values significantly reduce costs but at the expense of performance metrics. Notably, increasing the neighbor count $N$ from 3 to 5 leads to a counter-intuitive reduction in computational costs. This is attributed to our redundancy pruning, which discards a fusion candidate if the number of retrieved facts sharing the overlapping content reaches a majority threshold of $\lceil N/2 \rceil$. Under the $N=5$ setting (threshold of 3), this mechanism triggers more aggressive pruning of redundant information compared to $N=3$ (threshold of 2), thereby reducing overhead but slightly hampering accuracy due to over-filtering. Consequently, we adopted $\tau=0.85$ and $N=3$ as the balanced configuration for our main experiments, ensuring high performance within an acceptable computational budget.

\noindent\textbf{Impact of Retrieval Top-k.}
The quantity of retrieved memories (top-k) directly dictates the extent of contextual coverage and the noise level during inference. As illustrated in Figure~\ref{fig:topk_hyparam} and detailed in Appendix~\ref{appendix:topk}, as the retrieval count $k$ increases from 10 to 50, the method's average accuracy (ACC) exhibits a steady upward trend, rising from 69.54\% to 75.45\%. This performance gain is particularly pronounced in Single Hop tasks (+8.21\%), suggesting that broadening the retrieval scope provides more substantial evidentiary support for simple tasks. However, for Multi Hop and Temporal tasks involving complex reasoning, performance does not scale linearly with $k$. In fact, diminishing returns or even slight regressions are observed at $k=50$ (e.g., Temporal accuracy declines from 62.31\% at $k=10$ to 59.50\% at $k=50$). This phenomenon suggests that while expanding the retrieval window enhances the probability of capturing the correct answer, it concurrently introduces a significant volume of irrelevant distractor information; excessive redundancy can, counterintuitively, impede the LLMs' reasoning capabilities. Consequently, while a larger $k$ benefits overall metrics, handling specific complex tasks necessitates striking a delicate balance between retrieval breadth and information quality.

\noindent\textbf{Computational Complexity and Scalability.}
To comprehensively evaluate the computational overhead and operational efficiency of the AEG, we conducted a rigorous quantitative audit across the LoCoMo evaluation corpus, as detailed in Table~\ref{tab:component_efficiency_comparison}. By decoupling graph construction from inference, the offline fact extraction and event integration require 1,264 and 3,810 seconds respectively as one-time setup costs, so the online phase remains exceptionally lightweight. Specifically, vector retrieval for all 1,540 evaluation queries takes only 6.53 seconds in total. This negligible average latency of approximately 4 milliseconds per query, ensuring that the system maintains high responsiveness and practicality during continuous conversational streams.

Furthermore, the framework's memory footprint and graph expansion are tightly bounded by our redundancy pruning mechanism. As detailed in Table~\ref{tab:structural_stat}, across the evaluation corpus, the system actively identified and discarded 4,821 redundant fact clusters, and exceeding the 3,570 successfully integrated associative events in number. This high discard rate serves as a built-in saturation valve. As interactions grow lengthier and inevitably accumulate repetitive semantic patterns, the pruning mechanism aggressively throttles graph expansion, ensuring sustainable and highly efficient long-term memory operations.

\section{Conclusion}
In this paper, we introduced AnchorMem, a memory framework inspired by the Proust Phenomenon in cognitive theory, which uses atomic facts as retrieval anchors while preserving raw interaction as immutable context. By composing anchors into an associative event graph with higher-order event links, the framework binds multiple related facts into a shared event representation, enabling direct many-to-one integration of fragmented details into coherent events while avoiding noisy pairwise shortcuts. Experiments on LoCoMo across three open-source and closed-source model backbones show that AnchorMem consistently surpasses representative baselines, enabling models to leverage memories with practical efficiency in long-term interactions.

\section*{Limitations}
Although AnchorMem reduces the need for repeated rewriting by anchoring retrieval on atomic facts while preserving the original interaction history as immutable context, its effectiveness still depends on several design choices. Atomic fact extraction and fact to event grouping rely on model judgments, so ambiguous phrasing, implicit references, or inconsistent granularity can introduce missing or misgrouped anchors, and future work could explore alternative signals and connection mechanisms for linking facts more reliably. The current associative event graph is constructed with a single integration pass, and iterative multi-round integration may further refine event boundaries, and improve stability as interactions evolve.

\section*{Acknowledgments}
This work was supported in parts by Guangdong Basic and Applied Basic Research Foundation (2026A1515011358), Shenzhen Natural Science Foundation (JCYJ20250604181610014), and Intelligent Computing Center of Shenzhen University.

\bibliography{custom}

\clearpage  
\appendix
\onecolumn

\section{Experiment}
\label{appendix:exp}

\subsection{Benchmark and Baselines}
\noindent\textbf{Benchmark.} The LoCoMo~\citep{maharana-etal-2024-evaluating-locomo} benchmark is designed to evaluate the long-term memory capabilities of large language models. This dataset contains highly extended dialogues with each conversation having an average length of roughly 600 turns and 16k tokens. To comprehensively test memory performance the benchmark includes questions across four specific categories which are Single Hop, Multi Hop, Open Domain and Temporal.

\noindent\textbf{Baselines.} HippoRAG 2~\citep{RW-GR-gutierrez2025from} integrates a knowledge graph with Personalized PageRank. By identifying relevant triples to guide dense retrieval reranking, it enhances the discovery of Multi Hop relationships. Mem0~\citep{chhikara2025mem0} uses a hybrid storage mechanism of vector search and graph structures. It maintains a global state by dynamically extracting entities and user preferences from history to provide contextual guidance. A-Mem~\citep{RW-MS-xu2025amem} draws inspiration from the Zettelkasten method, transforming interactions into structured notes. It utilizes a memory evolution mechanism to dynamically update attributes and strengthen links between segments. MemoryOS~\citep{kang-etal-2025-memoryos} adopts operating system principles to construct three-tier memory hierarchy. It manages information migration across tiers using promotion strategies like dialogue-chain-based FIFO. LightMem~\citep{RW-MS-fang2025lightmem} is based on cognitive memory models, dividing processing into sensory, short-term, and long-term stages. It employs lightweight compression to filter redundancy and generates topic-aware summaries to optimize retrieval efficiency.

\subsection{Detailed Analysis of AEG Hyperparameters.}
\label{appendix:hyper_aeg}
In this section, we provide a granular sensitivity analysis of the AEG construction parameters, specifically the inter-fact similarity threshold $\tau$ (ranging from 0.70 to 0.95) and the number of retrieved neighbors $N$ ($N \in \{3, 5\}$). As detailed in Figure~\ref{tab:hyperpa_ablation}, the empirical data quantifies the tension between semantic preservation and system efficiency. We observe that relaxing the similarity constraint (e.g., $\tau=0.70$) encourages a denser fusion of facts. While this strategy maximizes information capture—peaking at an Accuracy of 72.73\%, it incurs substantial overhead in terms of runtime and update token usage. In contrast, tightening the threshold acts as a rigorous filter; for instance, the setting of $\tau=0.95$ with $N=5$ drastically minimizes consumption to merely 131.2 tokens. However, this aggressive pruning disconnects valid semantic links, leading to a demonstrable degradation in both Accuracy and F1 scores.

Interestingly, the numerical results reveal an inverse relationship between the neighbor number $N$ and resource consumption. Specifically, increasing $N$ from 3 to 5 results in a sharp decline in computational costs rather than an increase. For example, at $\tau=0.90$, token consumption drops precipitously from 2997.0 ($N=3$) to 612.6 ($N=5$). This efficiency gain validates our majority-based pruning mechanism: with a threshold of $\lceil N/2 \rceil$, the condition for discarding redundant fusion groups becomes easier to trigger at $N=5$ (threshold of 3) compared to $N=3$ (threshold of 2). Consequently, the $N=5$ setting intercepts a larger volume of fusion requests, saving resources but causing a slight accuracy regression due to over-filtering. Based on these comprehensive metrics, we identified the combination of $\tau=0.85$ and $N=3$ as the optimal operating point. It sustains a robust Accuracy of 69.54\% while maintaining manageable computational costs, thereby serving as the standard configuration for our main experiments.

\subsection{Detailed Data on Retrieval Top-k.}
\label{appendix:topk}
Table~\ref{tab:topk_detailed} presents the detailed numerical results regarding the impact of retrieval top-k. These specific values correspond to the trends visualized in Figure~\ref{fig:topk_hyparam} in the main text, breaking down the performance across different task types as top-k varies.

\begin{table*}[t!]
    \centering

    \renewcommand{\arraystretch}{1.2} 
    \setlength{\extrarowheight}{0pt}
    \newcommand{\mc}[1]{\multicolumn{1}{c}{#1}}
    \scalebox{0.68}{
    \begin{tabular}{cc ccc ccc}
        \hline
        \multicolumn{2}{c}{\textbf{Hyperparam.}} & \multicolumn{3}{c}{\textbf{Average}} &\multicolumn{1}{c}{\multirow{2}{*}{\textbf{Update Token (k)}}} &\multicolumn{1}{c}{\multirow{2}{*}{\textbf{Calls}}} &\multicolumn{1}{c}{\multirow{2}{*}{\textbf{Runtime (s)}}}\\
        
        \cmidrule(lr){1-2} \cmidrule(lr){3-5} 
        \mc{$\tau$} &\mc{$N$} & \mc{F1} & \mc{BLEU} & \mc{ACC} &\multicolumn{1}{c}{} &\multicolumn{1}{c}{} &\multicolumn{1}{c}{} \\
        \midrule
        0.70 &3 &52.68 & 44.44 &72.73 &13867.3 &11387 &8383.6 \\
        0.80 &3 &51.66 &43.64 &70.84 &8326.3 &8398 &5292.6\\
        0.85 &3 &50.57 &42.54 &69.54 &5129.1 &6577 &5073.7\\
        0.90 &3 &50.40 &42.40 &68.12 &2997.0 &5277 &2689.0\\
        0.95 &3 &49.79 &41.88 &68.18 &1571.3 &4244 &1933.5\\
        0.70 &5 &52.35 &44.22 &71.49 &13441.3 &9523 &8509.9\\
        0.80 &5 &51.11 &43.34 &69.61 &5636.4 &5852 &4189.9\\
        0.85 &5 &50.36 &42.45 &67.86 &2258.3 &4210 &2290.5\\
        0.90 &5 &50.22 &42.80 &67.92 &612.6 &3370 &1560.7\\
        0.95 &5 &49.42 &41.92 &66.95 &131.2 &3092 &1338.8\\
        
        \bottomrule
    \end{tabular}}
\caption{Impact of the inter-fact similarity threshold $\tau$ and the number of fact neighbors $N$ on performance and efficiency with Qwen2.5-32B-Instruct.}

\label{tab:hyperpa_ablation}
\end{table*}

\begin{table*}[t!]
    \centering
    \renewcommand{\arraystretch}{1.2} 
    \setlength{\extrarowheight}{0pt}
    \newcommand{\mc}[1]{\multicolumn{1}{c}{#1}}
    \scalebox{0.68}{
    \begin{tabular}{l  ccc  ccc  ccc  ccc  ccc}
        \hline
        \multicolumn{1}{c}{\multirow{2}{*}{\textbf{Top-k}}}  & \multicolumn{3}{c}{\textbf{Single Hop}} & \multicolumn{3}{c}{\textbf{Multi Hop}} & \multicolumn{3}{c}{\textbf{Open Domain}} & \multicolumn{3}{c}{\textbf{Temporal}} & \multicolumn{3}{c}{\textbf{Average}}\\
        
        \cmidrule(lr){2-4} \cmidrule(lr){5-7} \cmidrule(lr){8-10} \cmidrule(lr){11-13} \cmidrule{13-16}
        
        \multicolumn{1}{c}{} & \mc{F1} & \mc{BLEU} & \mc{ACC} & \mc{F1} & \mc{BLEU} & \mc{ACC} & \mc{F1} & \mc{BLEU} & \mc{ACC} & \mc{F1} & \mc{BLEU} & \mc{ACC} & \mc{F1} & \mc{BLEU} & \mc{ACC} \\
        \midrule
        10 &57.82 &49.76 &78.83  &35.53&25.71&58.51 &25.62&19.95&44.79 &53.12&45.26&62.31 &50.57&42.54&69.54 \\
        20 &61.20&52.17&83.12 &37.40&28.89&64.18 &24.35&18.47&42.71 &54.43&46.51&61.37 &53.1&44.63&72.40 \\
        30 &63.02&53.97&85.26 &40.62&32.13&68.09 &24.42&18.77&54.83 &54.76&46.41&59.91 &54.79&46.20&74.35 \\
        40 &63.47&54.42&85.37 &41.38&33.22&68.43 &25.49&19.70&47.92 &55.40&47.24&59.50 &55.37&46.88&74.55 \\
        50 &65.57&55.60&87.04 &40.84&33.68&68.09 &26.98&21.57&48.96 &54.58&46.26&59.50 &55.80&47.52&75.45 \\
        
        \bottomrule
    \end{tabular}}
\caption{Impact of the memory retrieval parameter top-k across different task categories with  Qwen2.5-32B-Instruct.}
\label{tab:topk_detailed}
\end{table*}

\section{Prompts}
\label{appendix:all_prompt}

\subsection{Prompt for Extracting Facts}
\begin{promptbox}{Prompt for Extracting Facts}
\label{prompt:ext_fact}
\small
-Goal-\\
You are an expert in extracting factual memories.\\

\# TASK

Your task is to extract memories from the snippets of dialogue between two speakers. This means identifying what each speaker would plausibly remember, including their own experiences, thoughts, plans, or relevant statements and actions made by others that impacted or were acknowledged by the speaker.\\

\# FOCUS DUAL SPEAKER

You must extract facts and memories for BOTH speakers involved in the conversation. Ensure the output list contains a comprehensive representation of both speakers' perspectives.\\

\# INSTRUCTIONS

1. Identify information that reflects speaker's experiences, beliefs, concerns, decisions, plans, or reactions — including meaningful input from one speaker that the other acknowledged or responded to.

2. Resolve all person, and event references clearly:

   - Include specific locations if mentioned.
   
   - Resolve all pronouns, aliases, and ambiguous references into full names or identities.
   
   - Disambiguate people with the same name if applicable.
   
3. Do not omit any information that speakers is likely to remember.

   - Include all key experiences, thoughts, emotional responses, and plans — even if they seem minor.
   
   - Prioritize completeness and fidelity over conciseness.
   
   - Do not generalize or skip details that could be personally meaningful to speaker.
   
4. Every memory MUST start with the Name of the speaker.

5. Output Format:

   - Return ONLY a valid JSON list of strings.\\

\{Example\}

\end{promptbox}

\subsection{Prompt for Integration Event}
\begin{promptbox}{Prompt for Integration Event}
\label{prompt:inte_event}
\small
-Goal-\\
You are a memory organization expert and storyteller.\\

\# TASK\\
Your task is to organize a comprehensive, detailed event Narrative by integrating information from multiple dialogue fragments. You will be provided with a set of ``Focus Topics'' (key facts extracted previously) and ``Source Contexts'' (raw dialogues containing the details).\\

Must extract memories from the snippets of dialogue between two speakers. This means identifying what each speaker would plausibly remember — including their own experiences, thoughts, plans, or relevant statements and actions made by others that impacted or were acknowledged by the speaker.\\

\# FOCUS DUAL SPEAKER\\
You must extract facts and memories for BOTH speakers involved in the conversation. Ensure the output list contains a comprehensive representation of both speakers' perspectives.\\

\# INSTRUCTIONS\\
1. Identify information that reflects speaker's experiences, beliefs, concerns, decisions, plans, or reactions — including meaningful input from one speaker that the other acknowledged or responded to.\\
2. Focus on ``Focus Topics'': Ensure all information related to the provided Focus Topics is included. Use the Source Contexts to fill in the why, how, where, and who regarding these topics.\\

3. Resolve all person, and event references clearly:

- Include specific locations if mentioned.

- Resolve all pronouns, aliases, and ambiguous references into full names or identities.

- Disambiguate people with the same name if applicable.

4. Preserve Details:

- Include specific locations, names, and objects.

- Capture the speakers' emotional responses (e.g., excitement, anxiety, gratitude) and their reasoning.

- Do not generalize specific details if they are relevant to the Focus Topics.

5. Third-Person Perspective: Write from an objective third-person perspective.

6. Output Format:

- Return ONLY a valid JSON list of strings.\\

\{Example\}
\end{promptbox}

\subsection{Prompt for LLM Judge}
\begin{promptbox}{Prompt for LLM Judge}
\label{prompt:llm_judge}
\small
Your task is to label an answer to a question as `CORRECT' or `WRONG'. You will be given the following data:\\
    (1) a question (posed by one user to another user), \\
    (2) a `gold' (ground truth) answer, \\
    (3) a generated answer\\
which you will score as CORRECT/WRONG.\\

The point of the question is to ask about something one user should know about the other user based on their prior conversations.\\

The gold answer will usually be a concise and short answer that includes the referenced topic, for example:\\
Question: Do you remember what I got the last time I went to Hawaii?\\
Gold answer: A shell necklace\\
The generated answer might be much longer, but you should be generous with your grading - as long as it touches on the same topic as the gold answer, it should be counted as CORRECT. \\

For time related questions, the gold answer will be a specific date, month, year, etc. The generated answer might be much longer or use relative time references (like ``last Tuesday'' or ``next month''), but you should be generous with your grading - as long as it refers to the same date or time period as the gold answer, it should be counted as CORRECT. Even if the format differs (e.g., ``May 7th'' vs ``7 May''), consider it CORRECT if it's the same date.\\

Now it's time for the real question:\\
Question: {question}\\
Gold answer: {gold\_answer}\\
Generated answer: {generated\_answer}\\

First, provide a short (one sentence) explanation of your reasoning, then finish with CORRECT or WRONG. 
Do NOT include both CORRECT and WRONG in your response, or it will break the evaluation script.\\

Just return the label CORRECT or WRONG in a json format with the key as ``label''.
\end{promptbox}

\section{Use of AI Assistants}
We acknowledge the use of Large Language Models (e.g., Gemini 3\footnote{\url{https://gemini.google.com/}}) to assist with language translation and stylistic polishing to improve the readability of this paper. Additionally, during the experimental design phase, we utilized LLMs to iteratively refine the prompts used in our proposed method, specifically to optimize their instruction-following capabilities and ensure strict adherence to the required output formats. The authors reviewed all AI-generated content and take full responsibility for the validity of the final manuscript.

\end{document}